\documentclass[conference]{IEEEtran}
\IEEEoverridecommandlockouts
\usepackage{cite}
\usepackage{amsmath,amssymb,amsfonts}
\usepackage{algorithmic}
\usepackage{graphicx}
\usepackage{textcomp}
\usepackage{xcolor}
\usepackage{subcaption}

\usepackage{todonotes}

\def\BibTeX{{\rm B\kern-.05em{\sc i\kern-.025em b}\kern-.08em
    T\kern-.1667em\lower.7ex\hbox{E}\kern-.125emX}}
\begin{document}

\title{Keke AI Competition:\\Solving puzzle levels in a dynamically changing mechanic space
}

\author{\IEEEauthorblockN{M Charity}
\IEEEauthorblockA{\textit{Game Innovation Lab} \\
\textit{New York University}\\
New York City, New York, United States \\
mlc761@nyu.edu}
% \and
% \IEEEauthorblockN{Sarah Chen}
% \IEEEauthorblockA{\textit{Independent}
% }
\and
\IEEEauthorblockN{Julian Togelius}
\IEEEauthorblockA{\textit{Game Innovation Lab} \\
\textit{New York University}\\
New York City, New York, United States \\
julian@togelius.com}
}
% \author{\IEEEauthorblockN{Anonymous}
% \IEEEauthorblockA{\textit{Anonymous Lab} \\
% \textit{Anonymous University}}
% }
% \and
% \IEEEauthorblockN{Sarah Chen}
% \IEEEauthorblockA{\textit{Independent}
% }

\maketitle

\begin{abstract}
% \todo[inline]{Abstract about the competition goes here}
The Keke AI Competition introduces an artificial agent competition for the game Baba is You - a Sokoban-like puzzle game where players can create rules that influence the mechanics of the game. Altering a rule can cause temporary or permanent effects for the rest of the level that could be part of the solution space. The nature of these dynamic rules and the deterministic aspect of the game creates a challenge for AI to adapt to a variety of mechanic combinations in order to solve a level. This paper describes the framework and evaluation metrics used to rank submitted agents and baseline results from sample tree search agents.
\end{abstract}

\begin{IEEEkeywords}
ai competition, sokoban, puzzle, baba is you, javascript
\end{IEEEkeywords}

\section{Introduction}
% \todo[inline]{Talk about why the competition is made and the benefits of developing bots capable of solving levels and games where the rules and mechanics of the game are constantly changing and can be modified by the player in the middle of the solution}

With the increasing depth and complexity of puzzle games comes the increasing need for intelligent solvers for these games. Most puzzle games have a set system of rules that can be followed and will never be changed at any point during play. For example, in the puzzle game Sokoban, a player must push each crate to designated positions on the map in order to solve the puzzle. However, the player is not allowed to pull these crates, thus creating a constraint for the player to avoid pushing these crates into corners or other dead-locked areas that they cannot push it out of and prevent completion of the puzzle. The push-pull mechanic of this game will never change nor be affected by the player's choices and sequential actions. These types of puzzles are straightforward and their domain knowledge of the problem stays consistent from level to level. Artificial intelligence (AI) solvers have been tested on such puzzle game domains, particularly to Sokoban because of its discretized, simplistic movement as well as its consistent rulespace, and successfully able to solve many levels of varying sizes and solution lengths. Such solvers typically use tree search algorithms, reinforcement learning, evolutionary algorithms, and rule-based algorithms to solve these levels \cite{yannakakis2018artificial}.

\begin{figure}[h]
 \centering
 \begin{subfigure}[h]{0.235\textwidth}
     \includegraphics[width=\textwidth]{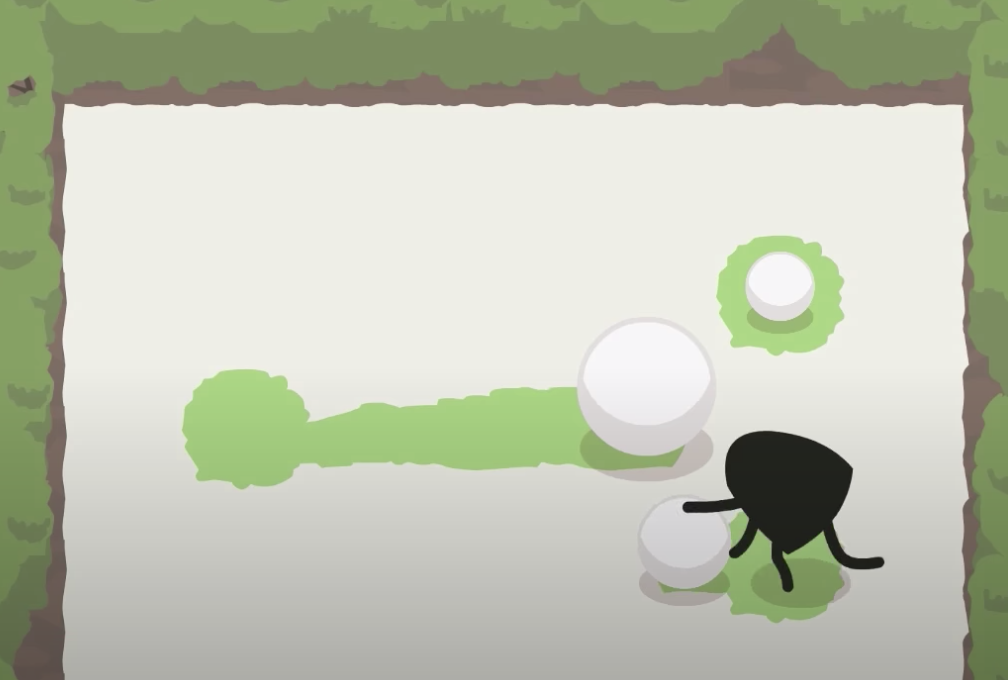}
     \caption{}
 \end{subfigure}
 \begin{subfigure}[h]{0.235\textwidth}
     \includegraphics[width=\textwidth]{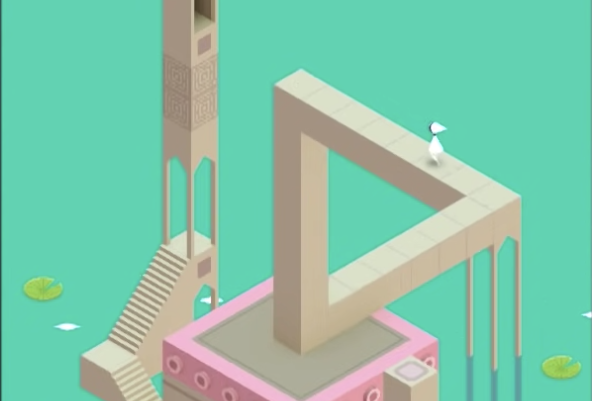}
     \caption{}
 \end{subfigure}
 \begin{subfigure}[h]{0.235\textwidth}
     \centering
     \includegraphics[width=\textwidth]{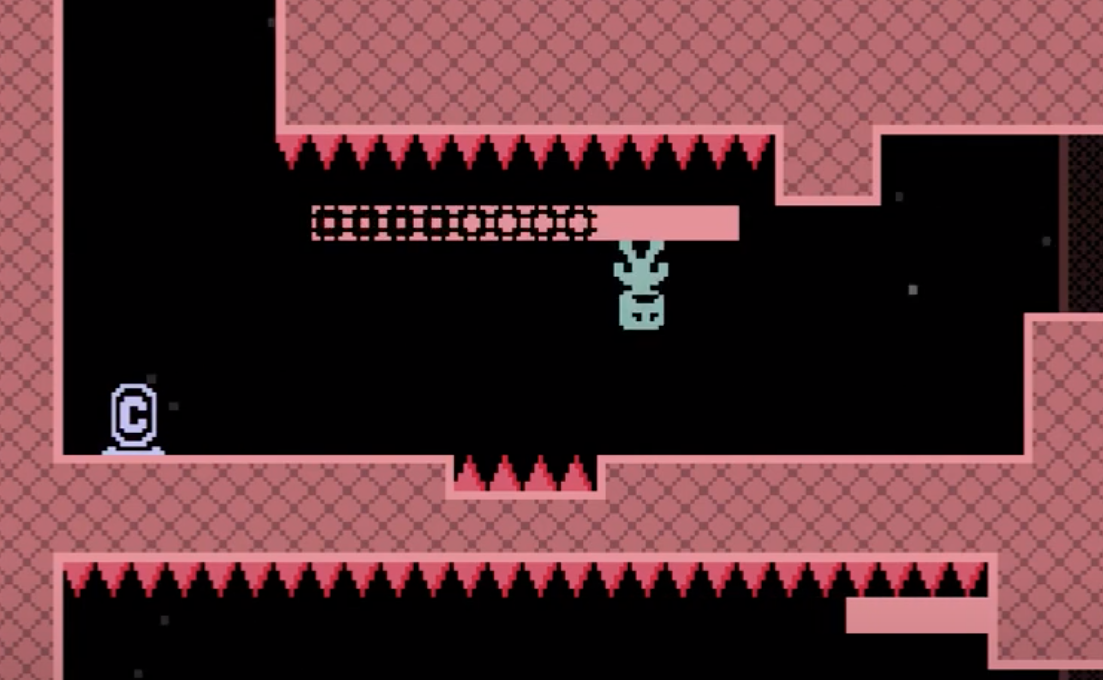}
     \caption{}
 \end{subfigure}
 \begin{subfigure}[h]{0.235\textwidth}
     \centering
     \includegraphics[width=\textwidth]{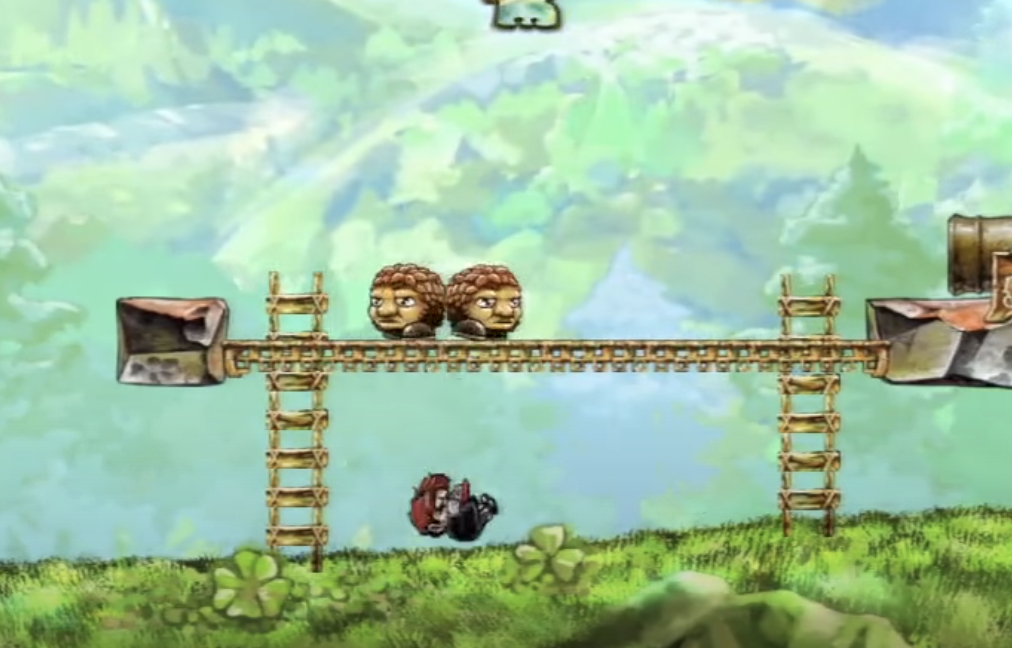}
     \caption{}
 \end{subfigure}
    \caption{Four games (clockwise starting top left: A Good Snowman is Hard To Build, Monument Valley, Braid, VVVVVV) with player-controlled dynamic mechanics that can temporarily or permanently affect the state of the game or player.}
\end{figure}

However, some games may have a dynamically changing mechanic space that are affected by the player's actions. With these types of games, the rule space is more dependent on when the rules or conditions are altered and become a part of the solution for the puzzle. These changes can cause permanent and irreversible effects to the game state, or be dependent on a certain sequence or combination of fluctuations in order to reach a particular game state. Some examples of this type of mechanic space can include resource changes such as in the game 'A Good Snowman is Hard to Build' (Draknek, 2015) where the amount of snow left on the ground is influenced by where the player decides to roll their snowball, perspective changes such as in Fez (Polytron Corporation, 2012) where the player has to change from 2D to 3D or where the player has to rotate the entire map such as in Captain Toad: Treasure Tracker (Nintendo, 2014) and Monument Valley (ustwo, 2014) in order to navigate through the level successfully, gravity changes such as in VVVVVVV (Terry Cavanagh, 2010) and Super Mario Galaxy (Nintendo, 2007) where the player flips the direction of gravity to navigate the level, or even time sequential changes such as in Braid (Number None, Inc., 2008) or Life is Strange (Dontnod Entertainment, 2015) where the player has to rewind time to complete certain objectives such as obtaining a key or specific information or destroying an enemy in order to progress forward. As such, dynamic mechanic spaces increases the complexity of the solution and requires more planning and deliberate decisions from the player in order to solve successfully. 

Baba is You (Hempuli, 2019) is one such dynamic puzzle game, where the player can alter the placement of sprites - similar to Sokoban. However, these placements of sprites can create rules that can influence the interaction and state conditions of the entire level, either temporarily or permanently. Such particular sprite placements and combinations are typically necessary to alter in order to solve the level. Because of the dynamically changing rules and sprite relationships, Baba is You is a multi-faceted challenge for AI solvers. The complexity of a level can vary because of these ever-changing rules; some rule combinations create drastic changes to the game states that would otherwise be unsolvable by even the best Sokoban solver agents. Baba is You is arguably more dynamic than the other competitions' domain games, as the objects in the level that a player interacts with can alter the entire state of the level itself. With the Baba is Y'all editor and database, the mechanic space and complexity of these levels can be explored in even more depth \cite{charity2022baba}. However, the AI solver used as an aide to validate the levels for solvability was lacking in performance, and could not solve levels that most human users could solve relatively quickly on their own. Thus, we would like to incorporate a better solver agent that can be more helpful for the user and possibly be capable of solving PCG levels offline without any user input. The Keke AI Competition invites entrants to submit their own solver agents for the Baba is You game that are capable of solving puzzles with a dynamically changing rule-base.

\section{Related Work}
% \todo[inline]{Talk about other competitions particularly related to puzzle/grid-world games or games where an agent might need to adapt to a different strategy during gameplay (GVG-AI competition, Hanabi competition, Bomberland competition, Nethack competition, Fighting game AI Competition)}

While Baba is You is not exactly like Sokoban - if anything, Baba is You could be considered a superset of Sokoban. There are many mechanic overlaps and the general premise of moving objects (whether crates or literal word block sprites) to specific positions (to solve the puzzle requirements itself or in order to create rules for altering the game state and interactions) maintains the same across games and are somewhat transferable. As such, many methods have been applied to solve Sokoban levels, including abstraction \cite{botea2002using}, dynamic programming using Prolog \cite{zhou2013tabled}, heuristic calculations on pattern databases with domain knowledge \cite{pereira2015optimal}, hyperspace graphing \cite{cook2019hyperstate}, automated curriculum planning with reinforcement learning \cite{feng2020novel}, forward-backward reinforcement learning \cite{shoham2021solving}, and Monte-Carlo tree search with reversibility compression \cite{cook2021monte}. But because of the dynamic nature of the mechanics in Baba is You, these solvers would not take into account how their actions may cause permanent alterations to the level and may not be as effective at finding a solution to the puzzle. Hence the need for a new application of solvers to a similar but more complicated domain. Bulitko and Ninomiya provide some insight to pathfinding and heuristic search in dynamic environments in both real-time and changing domains \cite{bulitko2011real} and domains with constraints \cite{ninomiya2015planning} - both relevant to Baba is You's environment. These methods could be applied along with the previous proposals of Sokoban solvers to create an adaptive planning agent capable of solving the adaptive levels.

Adaptive agents have been explored in previous years by other game AI competitions. The Hanabi competition is a cooperative multi-agent AI competition that uses the card game Hanabi as a domain. Agents take turns communicating information about other agents' hands (which they cannot see) and have to strategize over how much information to tell, what to communicate, and when information should be given in order to win the game and recieve as many points as possible from successfully playing cards. Because the playstyle and depth of interpretability of the other agents is unknown, agents have to adapt to their partners and the current state of the game in order to know which information is essential and how it would be percieved by others \cite{canaan2018evolving}. In constrast, the Fighting Game AI Competition is a real-time action game based on 2d arcade-style fighters. Agents have to survive and attack one other opponent in the game and have to time their attacks and defensive manuevers in order to succeed. The actions of the opponent player are immediate, so predicting and reacting to another agent's attacks must be done almost instantaneously \cite{lu2013fighting}. 

Grid-based domains have thoroughly been explored in games and AI - particularly with Sturtevant's work for general pathfinding done as proof for developing a benchmark \cite{sturtevant2012benchmarks} and as a competition \cite{sturtevant2015grid}. Previous game AI competitions have also been held that include grid-world based movement like the Keke AI Competition. These competitions also require adaptive strategies. The Bomberland AI Challenge is a real-time multi-agent competition where agents are able to move in 4 cardinal directions and must collect power-ups and strategically place bombs on the map in order to destroy their opponents. Agents cannot participate in the movement and actions of other agents on the map and have to adapt in real-time to both survive and attack \cite{zhang_2021}. The General Video Game AI (GVG-AI) Competition includes over 100 games all in a grid-world turn-based environment where participants submit Java agents that attempt to win as many games as possible \cite{liu_gvgai,perez2016general}. Because all of the games have different mechanic sets, sprite and level designs, and win conditions, an adaptive agent is needed to successfully play multiple games. The Nethack Learning Environment competition uses the roguelike dungeon-crawler turn-based game Nethack as a testbed domain for artificial agents where they have to navigate through multiple randomly generated levels of a dungeon and survive \cite{kuttler2020nethack}. This competition requires more generic and adaptive performing agents, as the game never generates the same rooms, encounters, or even effects of interaction. 

While there may be random events or unpredictable interactions from either the system or agent in Bomberland and Nethack, they stay within the scope of the interaction and the mechanics themselves never change. Where the Keke AI Competition differs from these competitions is how the mechanics of the game itself can be dynamically changed while the agent is solving and changing these rules may become part of the solution itself. A hazard or win condition can be created or destroyed with just a single move or arrangement of sprites. These changes stay deterministic as well, compared to the stochasticity of the other competitions, and can allow look-ahead strategies like the GVG-AI competition. An agent would not have to adapt to other players or the range of possibilities from a single action, but to how the mechanics and interactions of a level can be affected by a single move. 

\section{Baba is You}
\begin{figure}
    \centering
    \includegraphics[width=0.98\linewidth]{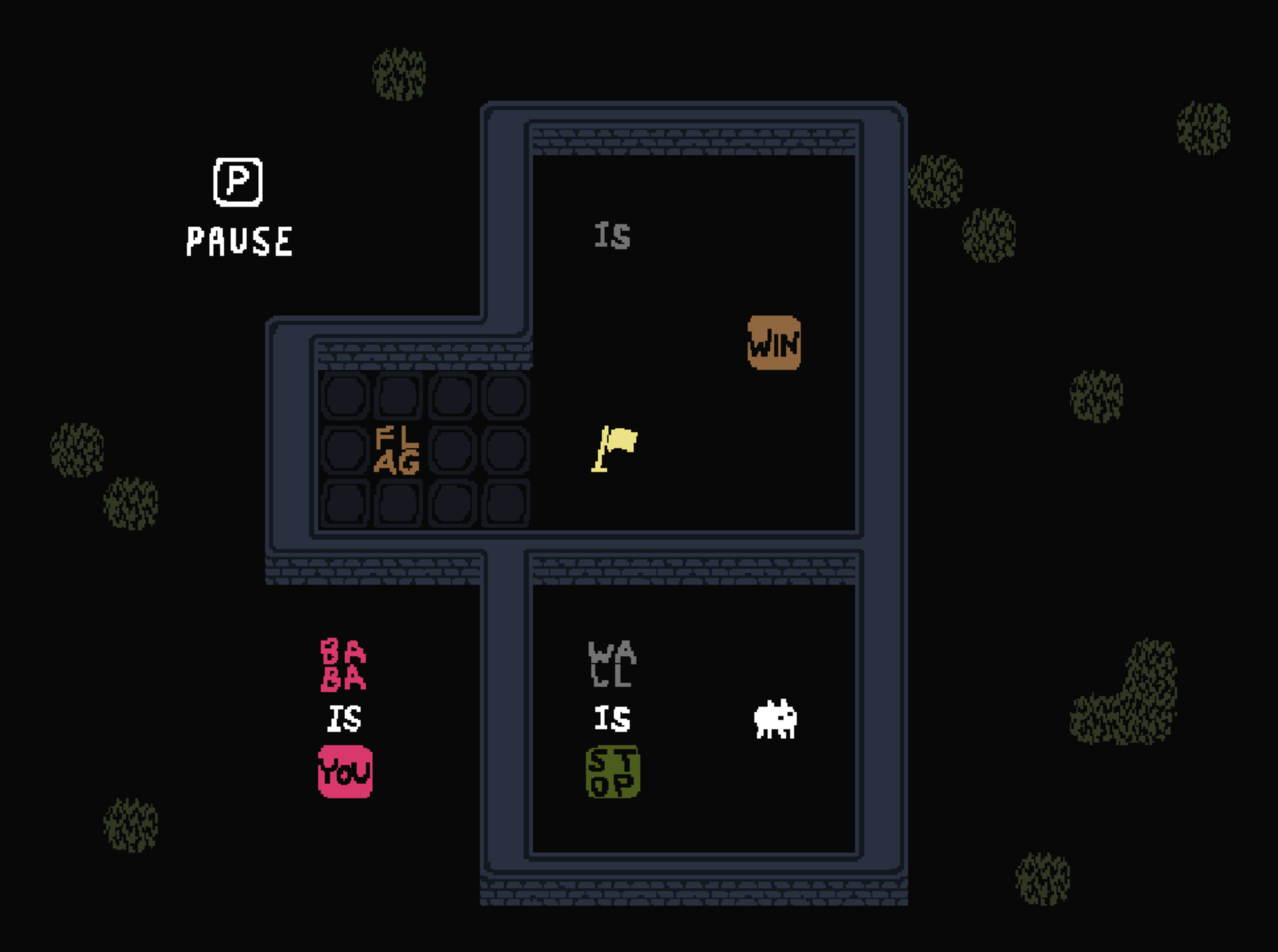}
    \caption{A level from Baba is You. Players have to break the active rule 'WALL-IS-STOP' and in order to push the word blocks on the other side together and create the rule 'FLAG-IS-WIN' in order to solve the level.}
    \label{fig:baba}
\end{figure}

Baba is You is a puzzle game developed by Arvi “Hempuli” Teikari originally for the 2017 Nordic Game Jam and then expanded to a full game with more mechanics.  Figure \ref{fig:baba} shows an example level from the fully released game. It has a similar gameplay style to Sokoban in that players must control a character and push sprite blocks around to solve the level. However, the rules for interacting with particular sprites on the level are changed by moving word tile blocks to form rules. Some rules can destroy sprites, prevent players from moving sprites, activate autonomously moving sprites, or even change the sprite the player currently controls. These rules of the level can be created or broken at any time, and much of the game involves manipulating the rules in a certain order or even specific orientation to allow the puzzle to be solved. 

\section{Keke AI Competition Framework}
% \todo[inline]{Talk about what went into the Keke Competition framework and the functions available to the agents for state evaluation as well as the outputs to said functions.}

The Keke AI Competition framework contains a simulator engine with state representation functions, a graphic interface evaluator that allows offline testing, and example baseline agents and level sets. Currently, the framework is written in JavaScript to work with the Baba is Y'all level editor web application. In future competitions, we would like to expand to allow for a Python framework so the competition could be more accessible and flexible for machine learning libraries such as PyTorch and Tensorflow.

\subsection{Simulation Engine}
% \todo[inline]{Talk briefly about the mixed initiative level editor for Baba is Y'all and tie back into the purpose of the competition to make better offline agent solvers for the tool}

The Keke AI simulation game used for the competition contains three major rule formats, where X is the name of sprite object (i.e. BABA, WALL, GRASS):
\begin{enumerate}
    \item \textbf{X-IS-(KEYWORD)} a property rule stating that all sprite instances of X have a certain property. These properties affect the interactions of other sprites (MOVE, STOP, HOT, COLD), win-conditions (WIN), lose-conditions (SINK, KILL), or player definitions and control (YOU) currently active in the level.
    \item \textbf{X-IS-Y} a transformative rule that changes all instances of sprites identified as X to the sprite Y. 
    \item \textbf{X-IS-X} a reflexive rule that prevents any transformations that occur on the sprite X. This differs from the previous rule X-IS-Y, as Y must be a sprite different from X in order for a transformation to occur. If a transformative rule is created, the X sprites will not transform into Y if the reflexive rule is active.
\end{enumerate}

The engine for the Keke AI Competition is the same engine used in the Baba is Y'all web application - a collaborative mixed-initiative level editor based off of the prototype version of the game Baba is You \cite{charity2020baba}. This engine contains 9 possible rule formats as shown in Table \ref{table:biy_rule_list}. We decided to use the same engine to allow for direct transferability of an agent to the level editor as a level solver. Currently, the website's current solver is lacking in performance, and cannot solve many levels - even simple ones that a user could solve in relatively little time. Ideally, we would like to use the highest performing agents from the competition in the level editor framework to optimize the level creation to submission pipeline and possibly allow for offline evaluation and procedurally generate levels for the database. The web application's ruleset has a much smaller percentage of rules than the fully released Baba is You game, as adding more rules could exponentially increase the complexity of a level and the potential mechanic rulespace found within the solution. However, we are open to expanding the ruleset to match the full game's in future years if the agents achieve a high threshold of solvability. 

The Keke AI Competition framework source code uses a state representation of the map similar to the Sokoban framework code by Khalifa \cite{khalifa2020multi}. The state itself contains an ASCII representation of the map, a list of active rules, a list of interactable and non-interactable objects, a list of word sprites (as these cannot be removed from the map), and lists of sprite objects assigned by attribute according to the currently active rules in the map. In other words, the states contain separated lists of active sprites that can be PUSHed, can KILL the player, can cause sprites to SINK, can MOVE, etc. Each state allows a look-ahead step that is passed one of 5 actions (up, down, left, right, or no movement) to allow for tree search agents to explore future states. The framework also includes state modification functions, such as copying a state, setting a state to a passed ASCII map's setup, reseting a state to the starting game map's setup, and more to allow for potential deep learning training. More detailed information about the states can be found on the framework's GitHub wiki page\footnote{https://github.com/MasterMilkX/KekeCompetition/wiki/Baba-Simulation-Code}. 

\begin{table}[t]
    \caption{Baba is Y'all Engine Rules}
    \centering
    \begin{tabular}{|p{0.2\linewidth}|p{0.7\linewidth}|}
    \hline
         Rule Type & Definition \\
    \hline
    \hline
        X-IS-X & sprites of class X cannot be changed to another class \\
        X-IS-Y & sprites of class X will transform to class Y \\
        X-IS-PUSH & X can be pushed \\
        X-IS-MOVE & X will autonomously move \\
        X-IS-STOP & X will prevent the player from passing through it\\
        X-IS-KILL & X will kill the player on contact\\
        X-IS-SINK & X will destroy any sprite on contact\\
        X-IS-[PAIR] & both rules 'X-IS-HOT' and 'X-IS-MELT' are present \\
        X,Y-IS-YOU & two distinct sprites classes are controlled by the player \\
    \hline
    \end{tabular}
    \label{table:biy_rule_list}
\end{table}

\subsection{Baseline Agents}
% \todo[inline]{Briefly describe the baseline agents included in the framework}

The Keke AI Competition framework contained 6 example agents that participants could use as a baseline to create their own agents. 3 of these agents were tree search agents that used the BFS, DFS, and the current solver agent used for the Baba is Y'all web application which uses a best-first search algorithm. One agent outputs a random sequence of 50 actions to the evaluator and another agent is provided as a blank template containing all of the necessary functions needed for evaluation. The last agent demonstrates an example of an agent that would not successfully pass the evaluation check upon submission. The user is free to modify these sample agents and include them in their own submission as a starting point.

\subsection{Offline evaluator interface}
% \todo[inline]{Talk about the process for agent evaluation and how users can develop their agents in JavaScript. How the user can upload their agents and evaluate them on the level test sets and see the various stats. Also talk about viewing the solution and GUI of the gameplay out.}
\begin{figure}
    \centering
    \includegraphics[width=0.98\linewidth]{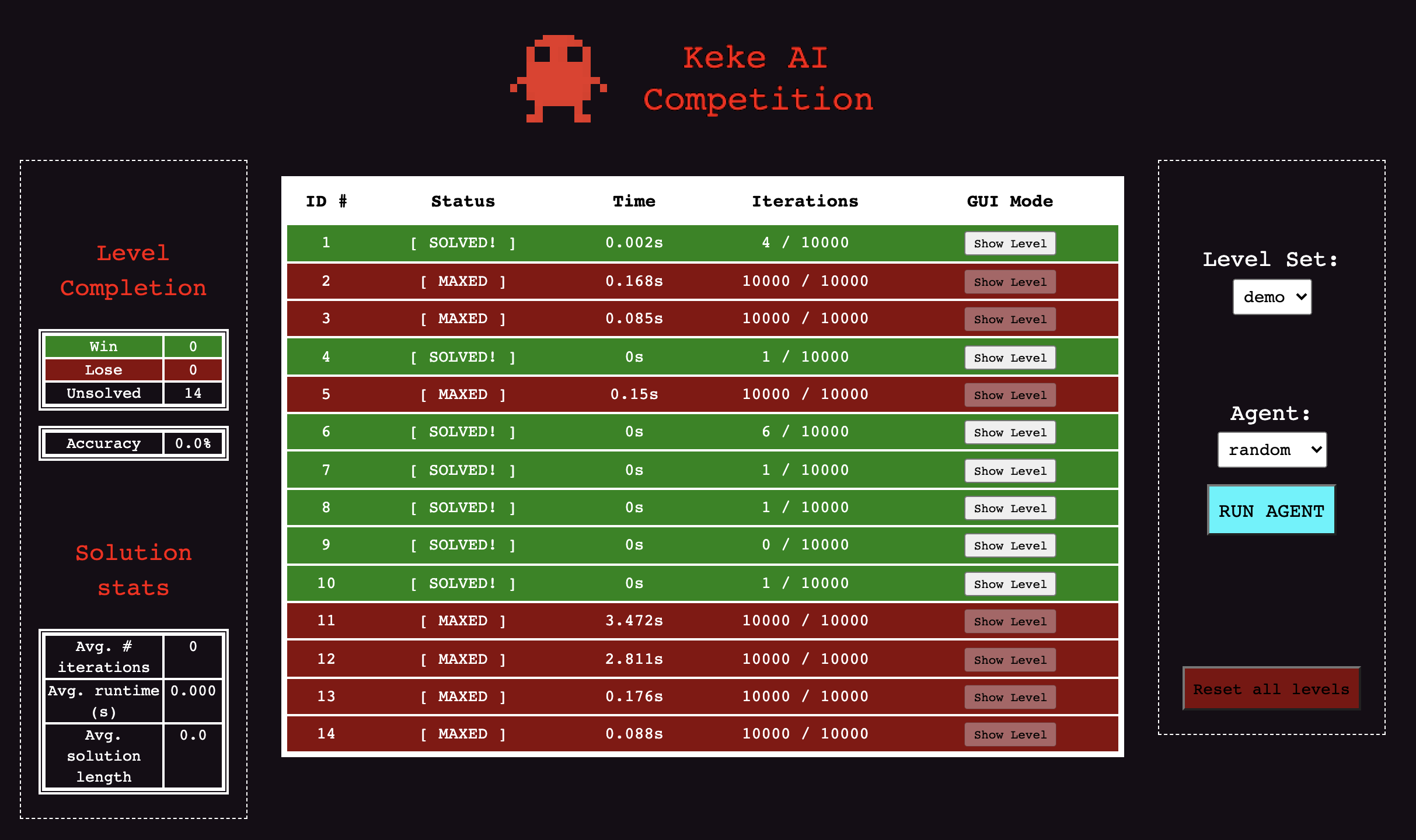}
    \caption{The evaluator screen of the offline Keke AI Competition offline interface}
    \label{fig:eval_mode}
\end{figure}

To test agents, the framework provides an offline web interface that allows users to test their agents on specific level sets as well as view the results of their agent's evaluations, statistical data for the agent on the level set, and view the agent's solution (or attempted solution) on a particular level in a rendering system. Figure \ref{fig:eval_mode} shows a screenshot of the offline interface with the indicators for solvability success, drop down menus for the agent and level set, and the statistics of the agent. We implemented this visual interface to provide more ease of access for the user and provide a lower barrier of entry to users who may not be as familiar with AI competitions in general. The interface also helps users identify quickly which levels were solved by the agent and how well they performed better than reading from a JSON report or a terminal output could. The interface can be viewed in any web browser (preferably Chrome) and users can select which levels to test on, however a text interface is also available via the terminal. 

\subsection{Level Sets and Baba is Y'all level extractor}
\begin{figure}
    \centering
    \begin{subfigure}[t]{0.4\linewidth}
        \centering
        \includegraphics[width=0.98\linewidth]{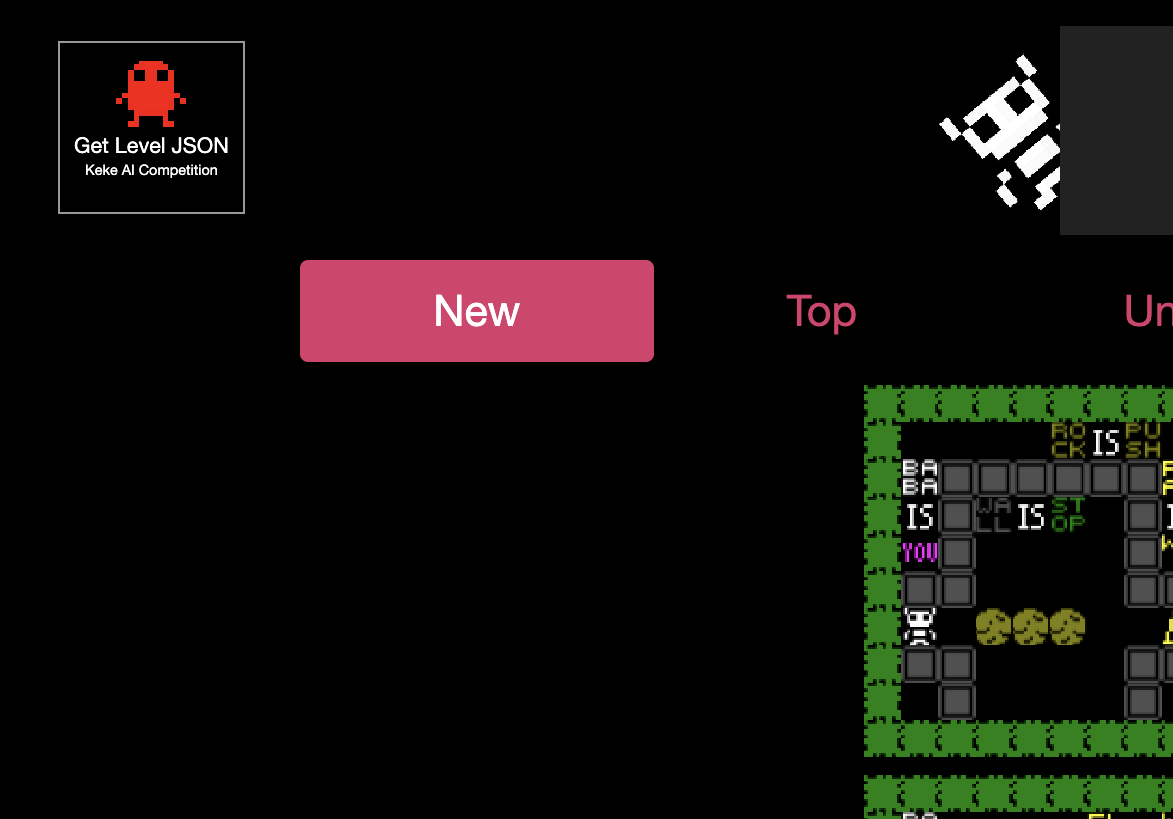}
    \end{subfigure}
    \begin{subfigure}[t]{0.4\linewidth}
        \centering
        \includegraphics[width=0.98\linewidth]{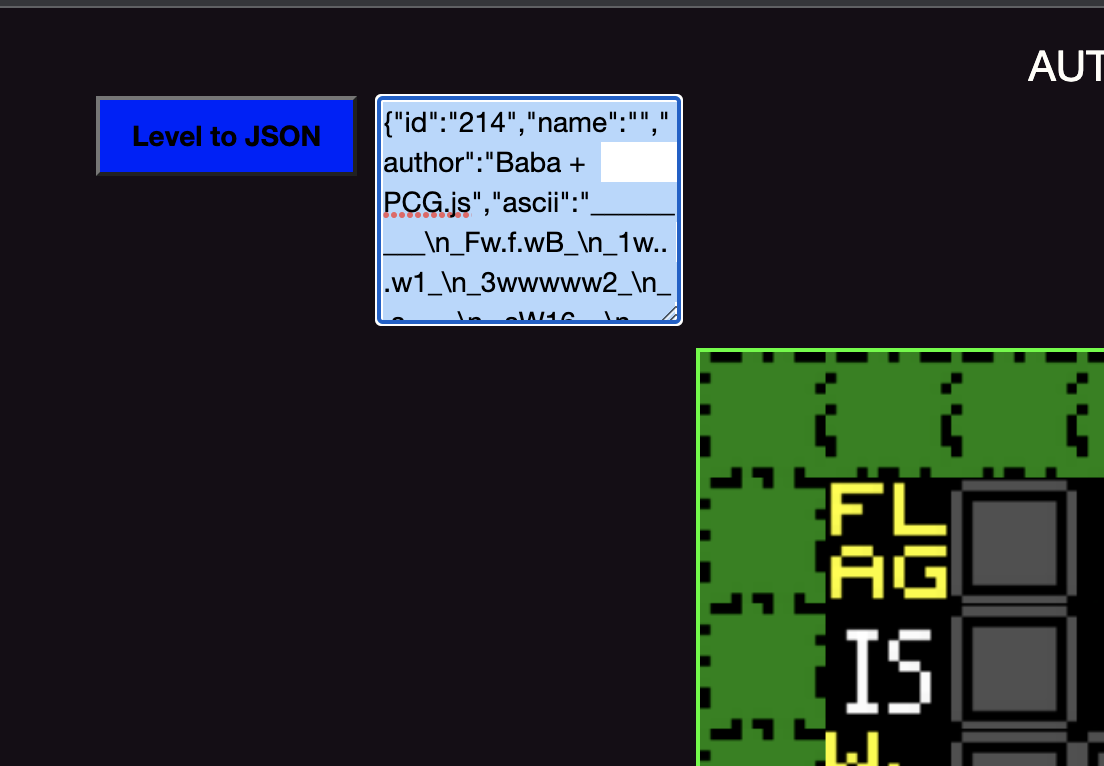}
    \end{subfigure}
    \caption{The Keke AI Competition export buttons on the main page (left) and the level testing screen (right)}
    \label{fig:biy_v2}
\end{figure}

% \todo[inline]{Talk briefly about the BiY level extractor so that users may create their own levels in the editor and save them for testing to a JSON or extract the currently uploaded levels for their own test set (via search, profile page, or the entire database)}
The level sets the agents are tested are made of JSON files containing ASCII levels created from the Baba is Y'all database. With this interaction between the Baba is Y'all level editor and the Keke AI framework, users can create their own training sets for their agents. The Baba is Y'all website was updated specifically for this competition to export level set JSONs in the framework's format. Users can extract levels from their own profiles, from a search result, or from the entire database (shown as a button in the top left of Figure \ref{fig:biy_v2}). The level tester screen of the Baba is Y'all site also allows users to export singular levels in the JSON format - whether they are submitted to the database or not - so that users can create their own level set files or add to any existing one.

% \begin{figure}
%     \centering
%     \includegraphics[width=0.9\linewidth]{imgs/new_baba_export.png}
%     \caption{A screenshot of the Baba is Y'all (v2) level editor web application}
%     \label{fig:biy_v2}
% \end{figure}

\section{Competition Website}
\begin{figure}
    \centering
    \includegraphics[width=0.98\linewidth]{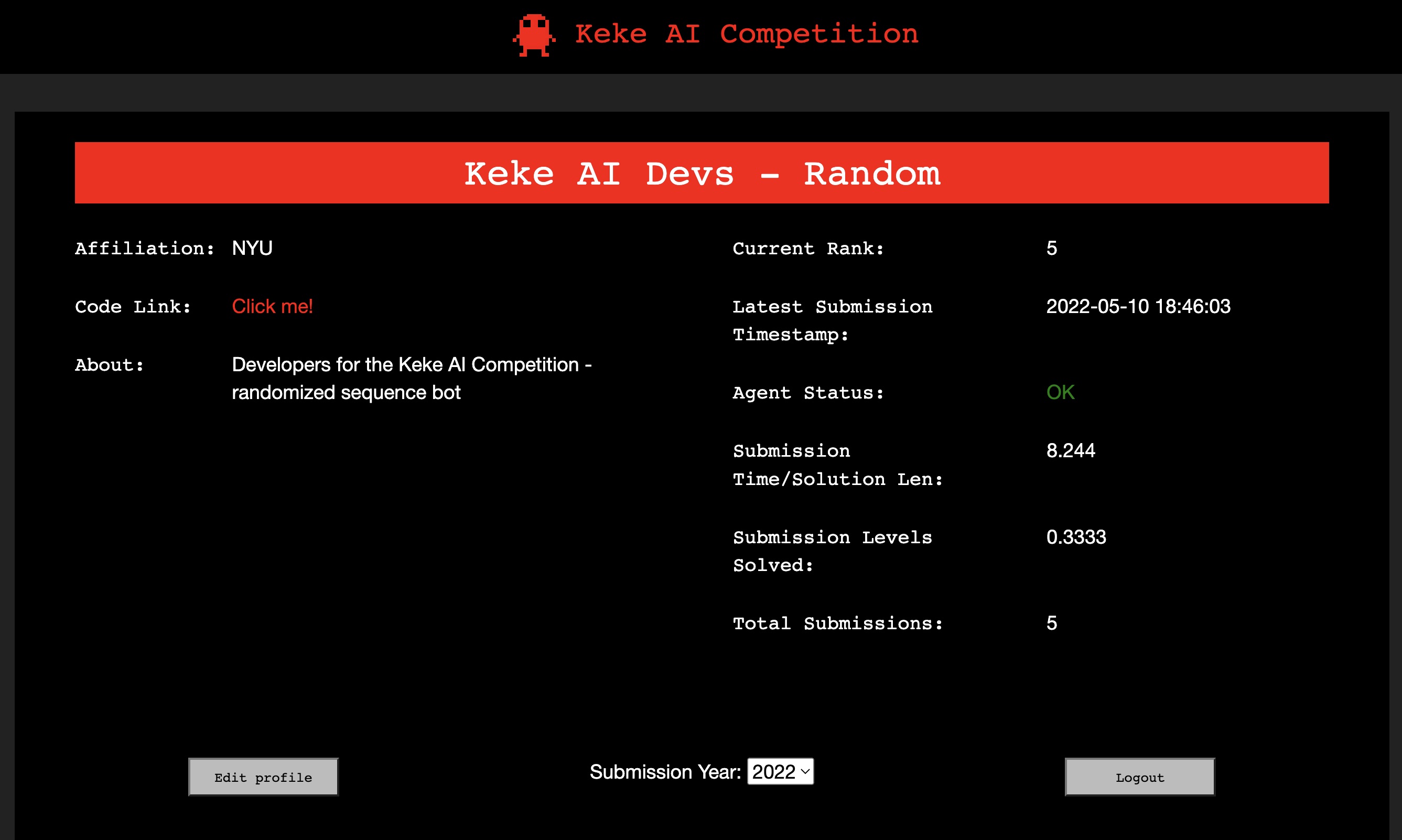}
    \caption{A screenshot of an example profile page on the Keke AI Competition website}
    \label{fig:profile}
\end{figure}

% \todo[inline]{Talk about how the registration process occurred for the competition and how new agents for each team could be submitted.}
This competition was developed using the guide written by Togelius on running a game AI competition \cite{togelius2014run}. Agents are submitted to the competition website
% \footnote{keke-ai-competition.com} 
hosted on an Amazon Web Services Ubuntu server and evaluated automatically by back-end scripts monitoring the website.
Participants  were required to register a team under a email-password login system and could provide further details and information about their team on the profile page. After registering their team, participants could submit their JavaScript agents and the server would evaluate them on a set of test levels. Figure \ref{fig:profile} shows an example profile page for the "Keke AI Devs" team and editable information fields. 

Users could navigate through the main screen of the website to find any necessary links they would need for the competition, including links to social media, the framework, the Baba is Y'all site, the Wiki, and a the walkthrough video for submitting an agent. The site also featured an about page with rules for submission.

\subsection{Competition evaluator}
% \todo[inline]{Describe how the competition agents are evaluated and under what basis. Talk about the levels and how they were designed with what challenges in mind (i.e. backtracking, mazes, etc)}
The website had a readily available leaderboard that showed each team's agent performance on the evaluation levels. These levels are not included on the original Baba is Y'all website and cannot be downloaded or explicitly trained on by participants. Teams are ranked on agent performance that is evaluated in the following order:
\begin{enumerate}
    \item \textit{Code error} - agents that error out in the preprocessing step are automatically placed at the bottom
    \item \textit{Average time\textsuperscript{-1} / solution length} (referred henceforth as timesteps) - agents that give shorter solutions solved in quicker time are ranked higher.
    \item \textit{Time submitted} - in the case of a tie in all other categories, the agent submitted earlier is ranked higher
\end{enumerate}
We designed the "average time\textsuperscript{-1} / solution length" as the evaluation metric for the competition to encourage participants to create agents that could find solutions to the puzzles in the fewest amount of steps and in the shortest amount of time possible. 

Evaluation of the agents is completely autonomous and handled by the website server. Once a team submits their agent, the JavaScript file is verified to have the required functions necessary to run the agent on the evaluation framework. After this check, the server will directly test the agent on the evaluation levels and save the results of the team's agent to the database to be passed to the leaderboard. The rankings are updated to reflect the relative scoring of the team's agent - and users can typically receive a status update within 20 minutes of submission. The status of the agent is also shown on their team profile page in case the agent had any errors while running. 

\subsection{Evaluation levels}

\begin{figure}[h]
 \centering
 \begin{subfigure}[h]{0.235\textwidth}
     \includegraphics[width=\textwidth]{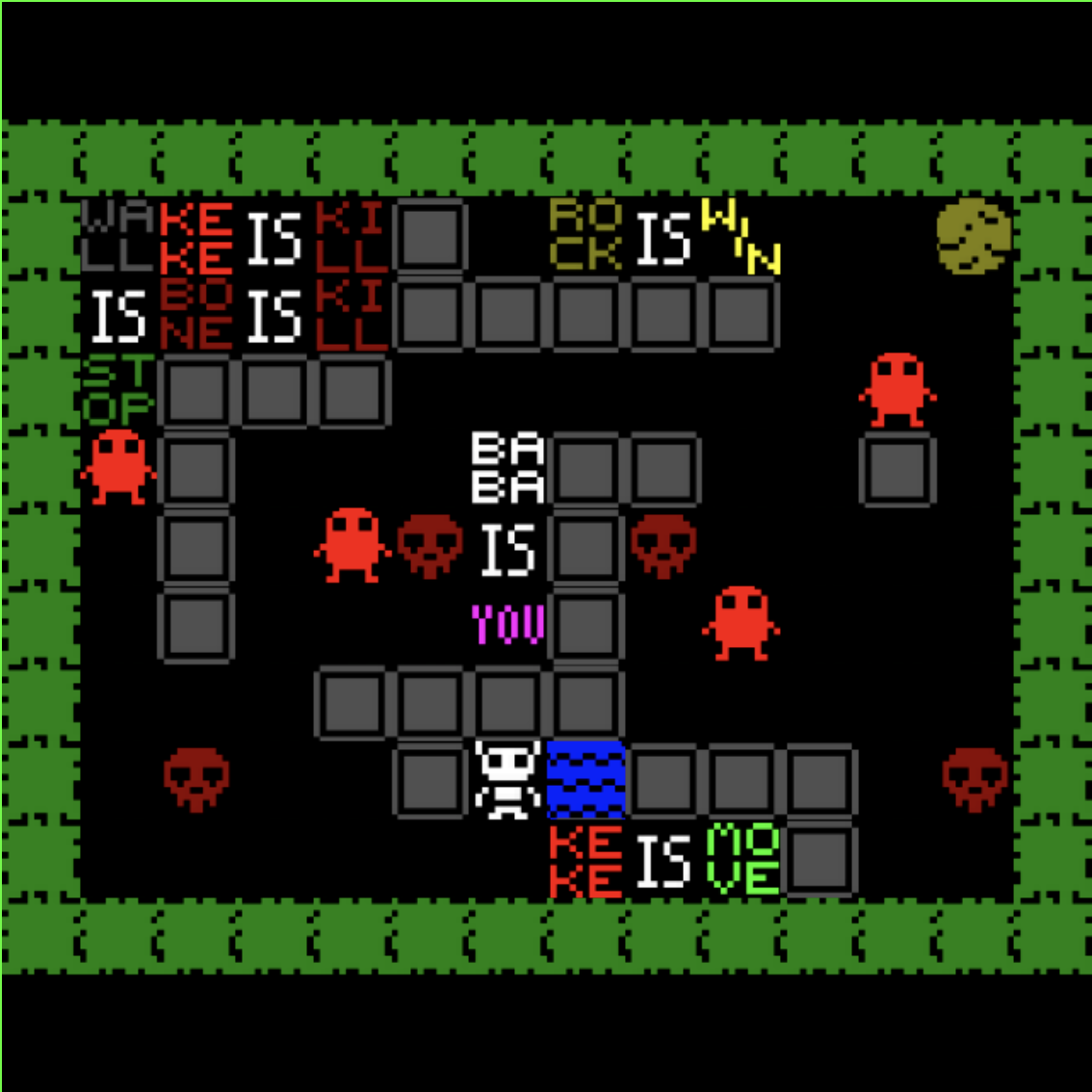}
     \caption{}
 \end{subfigure}
 \begin{subfigure}[h]{0.235\textwidth}
     \includegraphics[width=\textwidth]{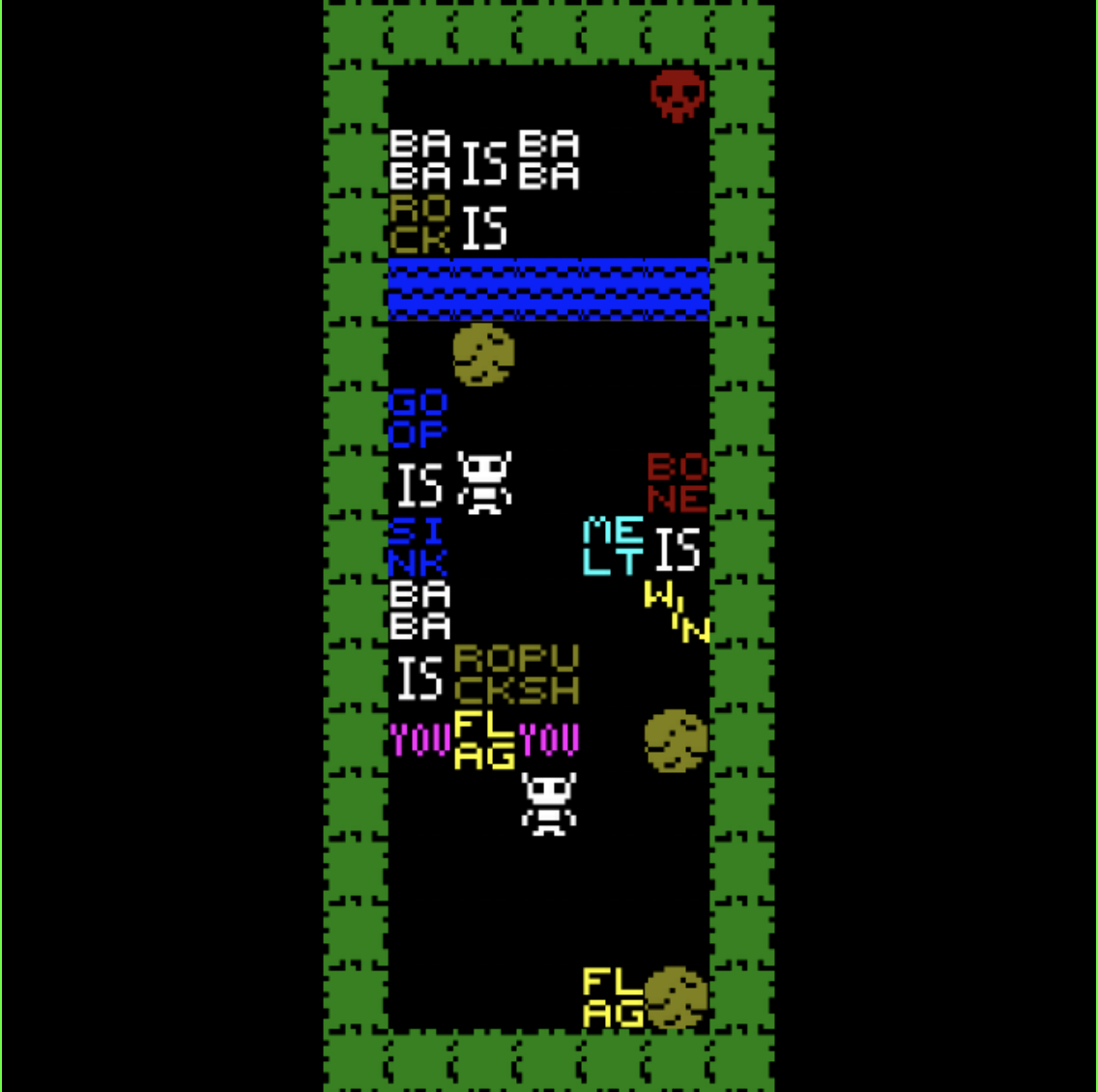}
     \caption{}
 \end{subfigure}
 \begin{subfigure}[h]{0.235\textwidth}
     \centering
     \includegraphics[width=\textwidth]{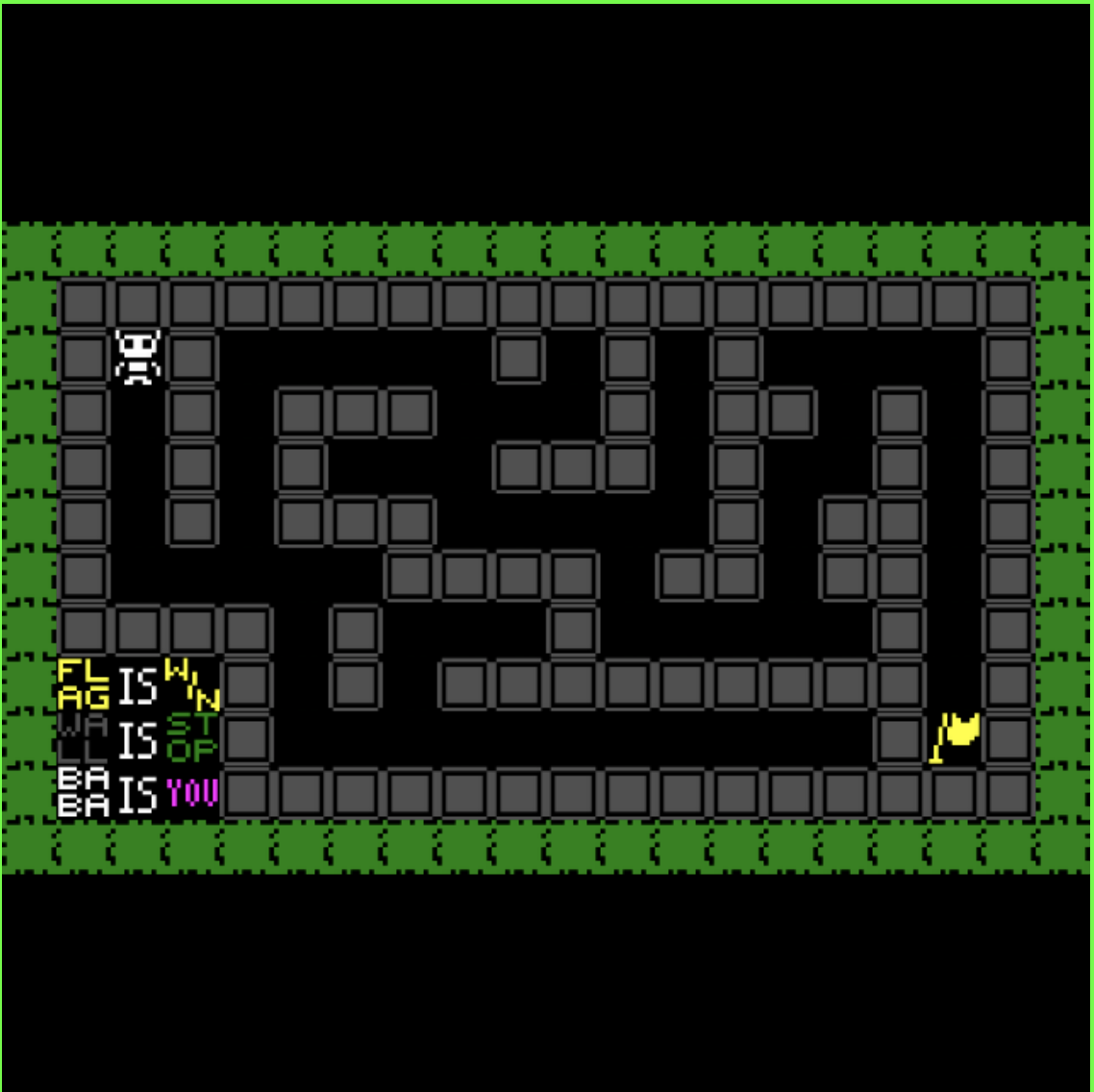}
     \caption{}
 \end{subfigure}
 \begin{subfigure}[h]{0.235\textwidth}
     \centering
     \includegraphics[width=\textwidth]{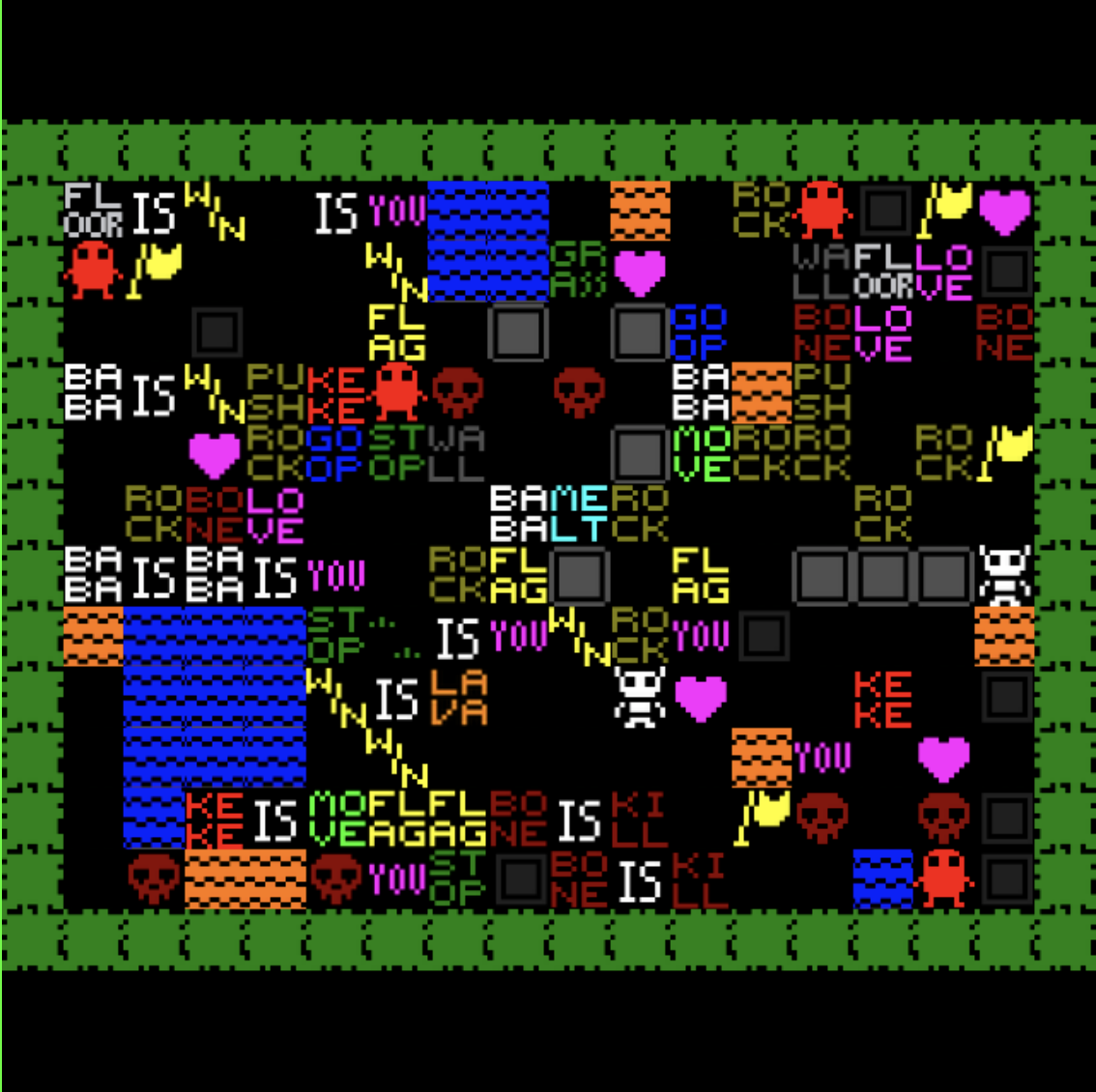}
     \caption{}
 \end{subfigure}
    \caption{Four levels from the first round evaluation set of the Keke AI Competition}
    \label{fig:eval_levels}
\end{figure}

For the first round of the competition, the levels made for evaluation were created with the intention to fool specific types of agents. Figure \ref{fig:eval_levels} shows 4 sample levels from the evaluation set. Level A is intended to require precise movements from the player by having a tunnel in order to successfully manuever around the autonomously moving hazard - a harder challenge for random sequence agents. Level B requires the manipulation of the word blocks in order to access the winnable sprite on the other side of a barrier - thus requiring an understanding of essential word blocks but also their secondary function as movable objects themselves. Level C is a standard maze level, which can be easily completed by a human, but provides more computation from a tree search bot and would be nearly impossible from a random sequence bot. Lastly, Level D is intended to overwhelm both bots and human players with the number of sprites, words, and rules active on the level and increase computation time, but can actually be solved in one move because the player sprite is already a win condition. 

There are more types of levels in the evaluation set that take advantage of the bots algorithms, such as a lock and key method, destroying sprites on a single move, requiring alteration of rules before moving to the win condition, and more. For future iterations of the competition we would like to design levels that exploit the computation requirement and decision processes of as many bots and algorithms as possible to facilitate a development direction towards generic solving.

\subsection{Baseline agent performance}

The leaderboard initially had the 4 baseline agents included the framework: the BFS-algorithm agent, the DFS-algorithm agent, the random sequence agent, and the default (Baba is Y'all solver) agent. After teams submit their agents, rankings would be re-ordered accordingly.

Out of the 4 baseline agents the default agent performed the best with 53.33\% of evaluation levels solved and average of 34.123 timesteps. In second and third ranking were the tree search agents, the BFS and DFS agents. The BFS agent had 40\% of levels solved at an average of 0.693 timesteps while the DFS agent had 33.33\% of levels solved at an average of 34.281 timesteps. Lastly, the random agent performed the 4th best, solving 33.33\% of levels - the same as the DFS agent - but had a lower average timestep of 8.244 and therefore was ranked lower. 

\section{Conclusion and Future Improvements}
% \todo[inline]{Talk about drawbacks and future improvements to the competition setup and the framework.}
Future iterations of the competition will include a Python framework to allow for more accessibility and a more familiar and widely used programming language for participants. Since JavaScript was not as well known or widely used in the AI and Games community as compared to Python, this could create a larger learning curve and deter participants from trying to develop their own agents. 

We would like to add more baseline agents to the framework as well, with a focus on creating agents that aren't necessarily tree-search based. Such algorithms we would like to use for the agents include evolutionary search (by mutating and evolving action sequences,) Monte-Carlo Tree Search, reinforcement learning, rule-based algorithms, and neuro-evolution agents. By providing a more diverse set of baseline agents, participants would be able to have alternative starting points for their own agents and be able to gauge the baseline agents' performance on the evaluated levels. 

At the conclusion of each year's competition, we would like to incorporate the winning agent into the Baba is Y'all website. We would also like to apply the agent, if a generalizable algorithm emerges from the winning agent, to other game domains with dynamically changing mechanics outside of Baba is You.

This competition provides a new domain of game design for artificial agents that involves dynamically changing rules for the game Baba is You. We hope this competition can allow new and innovative agents to emerge from this type of puzzle game and provide insight to the design process and evaluation of these agents in an adaptive game environment.

\section{Acknowledgements}
The authors would like to thank Sarah Chen (schen1337@gmail.com) for her contributions to the BFS and DFS agent codes.

\bibliography{ref}
\bibliographystyle{IEEEtran}

\end{document}